\documentclass[10pt,twocolumn,letterpaper]{article}

\usepackage{iccv}
\usepackage{times}
\usepackage{epsfig}
\usepackage{graphicx}
\usepackage{amsmath}
\usepackage{amssymb}
\usepackage{multirow}
\usepackage{subcaption}
\usepackage[pagebackref=true,breaklinks=true,letterpaper=true,colorlinks,bookmarks=false]{hyperref}
\usepackage{enumitem}
\setlist{nolistsep}
\usepackage{wrapfig}

\iccvfinalcopy % *** Uncomment this line for the final submission
\def\iccvPaperID{6950} % *** Enter the ICCV Paper ID here

\ificcvfinal\fi

\usepackage{color}
\definecolor{forest-green}{rgb}{0.55, 1.0, 0.55}
\definecolor{bubblegum}{rgb}{0.98, 0.85, 0.84}
\definecolor{new-blue}{rgb}{0.5, 0.5, 0.99}
\definecolor{darkorchid}{rgb}{0.6, 0.2, 0.8}
\usepackage{soul}
\newcommand{\model}{\texttt{ConTraCon}}

\usepackage[utf8]{inputenc} % allow utf-8 input
\usepackage[T1]{fontenc}    % use 8-bit T1 fonts
\usepackage{url}            % simple URL typesetting
\usepackage{booktabs}       % professional-quality tables
\usepackage{amsfonts}       % blackboard math symbols
\usepackage{nicefrac}       % compact symbols for 1/2, etc.
\usepackage{microtype}      % microtypography
\usepackage{xcolor}         % colors
\usepackage{float}

\makeatletter
\def\@fnsymbol#1{\ensuremath{\ifcase#1\or \dagger\or \ddagger\or
   \mathsection\or \mathparagraph\or \|\or **\or \dagger\dagger
   \or \ddagger\ddagger \else\@ctrerr\fi}}
    \makeatother

\begin{document}
\title{Exemplar-Free Continual Transformer with Convolutions}

\author{Anurag Roy$^{1}$ \ \ \ Vinay K. Verma$^{2,}$\thanks{Work started before joining Amazon} \ \ \ Sravan Voonna$^{1}$ \ \ \ Kripabandhu Ghosh$^{3}$ \ \ \ Saptarshi Ghosh$^{1}$ \ \ \ Abir Das$^{1}$\\
$^{1}$IIT Kharagpur, $^{2}$IML Amazon India, $^{3}$IISER Kolkata\\
\small{\texttt{\{anurag\_roy@,vg.sravan@,saptarshi@cse.,abir@cse.\}iitkgp.ac.in}}, \small{\texttt{\{vinayugc,kripa.ghosh\}@gmail.com}}
}
\maketitle

\begin{abstract}
Continual Learning (CL) involves training a machine learning model in a sequential manner to learn new information while retaining previously learned tasks without the presence of previous training data. Although there has been significant interest in CL, most recent CL approaches in computer vision have focused on convolutional architectures only. However, with the recent success of vision transformers, there is a need to explore their potential for CL. Although there have been some recent CL approaches for vision transformers, they either store training instances of previous tasks or require a task identifier during test time, which can be limiting. This paper proposes a new exemplar-free approach for class/task incremental learning called \model{}, which does not require task-id to be explicitly present during inference and avoids the need for storing previous training instances. The proposed approach leverages the transformer architecture and involves re-weighting the key, query, and value weights of the multi-head self-attention layers of a transformer trained on a similar task. The re-weighting is done using convolution, which enables the approach to maintain low parameter requirements per task. Additionally, an image augmentation-based entropic task identification approach is used to predict tasks without requiring task-ids during inference. Experiments on four benchmark datasets demonstrate that the proposed approach outperforms several competitive approaches while requiring fewer parameters.\footnote{Project Page: \url{https://cvir.github.io/projects/contracon}
}
\end{abstract}

\section{Introduction}

Humans excel at solving newer tasks without forgetting previous knowledge. Deep neural networks, however, face the challenge of forgetting old knowledge when trained for a novel task.
This problem, known as {\it Catastrophic Forgetting}, is a fundamental challenge
if a model needs to learn when data arrives in a sequence of non-overlapping tasks.
\emph{Continual Learning}~\cite{aljundi2017expert,TCS2020Hadsell,PNAS2017kirk,NN2019parisi,cvpr2017icarl,neurips2019rolnick} aims to handle such problems arising from non-stationary data to retain previously learned knowledge as well as acquire knowledge from new data with limited or no access to previous data.

\begin{figure}[t!]
    \centering
    \includegraphics[scale=0.2]{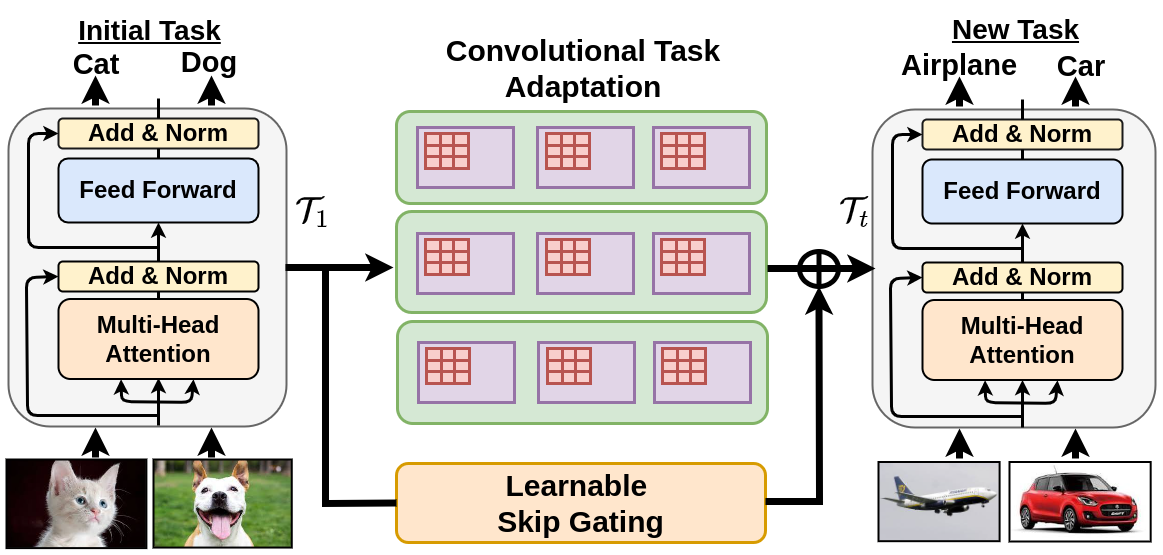}
    \caption{\small The proposed \model{} architecture. $T_1$ represents a transformer model trained on a previous task which is adapted to a new task using only a few learnable parameters and temporal skip-gating to a new transformer $T_t$ on the Right.
    }
    \label{fig:transformer_cl_fig1}
\end{figure}

Although continual learning in computer vision has witnessed remarkable progress, most of the methods are tailored for CNNs. 
Recently, vision transformers have shown promising results in big data regime~\cite{carion2020end,iclr2021vit,iccv2021swin}.
However, data hungry vision transformers and data-scarce continual learning do not seem to go hand in hand, and thus continual learning in vision transformers has received relatively little attention.
Early continual learning approaches on ConvNets relied on exemplar rehearsal which re-trains newer models on previous data instances stored in a fixed size memory buffer~\cite{cvpr2021rm,icpr2020ert,chaudhry2019continual,hayes2020remind,lopez2017gradient,prabhu2020gdumb,cvpr2017icarl,neurips2019rolnick}.
Dytox~\cite{douillard2021dytox} and LVT~\cite{wang2022lvt} are contemporary works addressing continual learning on vision transformers using such previously stored data.
However, storing task samples in raw format may not always  be feasible, especially for tasks where long-term storage of data is not permitted owing to privacy or data use legislation~\cite{he2022exemplar}.
{\it Dynamic architecture} based methods, on the other hand, grows an initial model dynamically or rearranges its internal structure,~\cite{fernando2017pathnet,serra2018overcoming, singh2020calibrating,iclr2018den} with the arrival of new tasks without the need for storing previous data.

The main focus of this paper is in modeling the new set of parameters for new tasks with as low overhead as possible.
We conjecture that better representation learning capability of a transformer keeps it ready for easily adapting to new tasks by \emph{manipulating} the already learned weights.
Convolution offers a cheap way to manipulate transformer weights and the weight matrices are natural fits to convolution as input and output.
In this paper, we propose a dynamically expandable architecture with novel convolutional adaption on previously learnt transformer weights to obtain new weights for new tasks.
In addition to this, we employ a learnable skip-gating which learns to convexly combine the old and the convoluted weights.
This helps significantly in maintaining the stability-plasticity tradeoff by balancing between how much to retain and how much to forget.
The resulting proposed approach -- \textbf{Con}tinual \textbf{Tra}nsformer with \textbf{Con}volutions (\model{}) (ref. Fig.~\ref{fig:transformer_cl_fig1}) -- not only leverages a vision transformer's learning capability but also does it with significantly low memory overhead per task as is shown in our experiments.

Specifically, for each new task, we reweigh the key, query, and value weights of the previously learned transformer by convolving them with small filters.
Transformers are known for their ability to capture long-range dependencies between patches.
Convolution, on the other hand, exploits a local neighborhood only.
Such local dependency not only restricts large changes in the weights for the new task from the old tasks, but also allows us to achieve this with very little increase in the model size.
As the convolution weights are separate for different tasks, during inference, this will require the information of which task an image belongs to.
However, in class incremental continual learning, images may come arbitrarily without associated task information.
To tackle the challenging scenario, we propose a novel entropy-based criterion to infer the task before calling for the corresponding task-specific convolution weights
In particular, our proposed approach creates multiple augmented views of a test image and evaluates the agreement of their predictions among different task-specific models.
The task-id of the image is determined by identifying the task giving the lowest entropy value of the average predictions from the various augmented views.
The task-specific model with the most consistent and confident predictions across different augmentations corresponds to the correct task for the image.

Extensive experiments on four benchmark datasets, considering both the availability and unavailability of task-ids at test time, demonstrate the superiority of our proposed method over several state-of-the-art methods including the extension of popular approaches on CNNs to transformers.
Despite being exemplar-free, our approach outperforms state-of-the-art exemplar-based continual learning approaches that use transformers as backbone architectures (Dytox~\cite{douillard2021dytox} and LVT~\cite{wang2022lvt}) with an average improvement of $5\%$.
Moreover, our approach accomplishes this with only about $\sim 60\%$ parameters used by above models.
We also perform ablation studies highlighting the contribution of each component in our approach.
To summarize, our key contributions include:
\begin{itemize}[leftmargin=*]
    \item We propose \model{}, a dynamic architecture for CL on transformers. We use convolution on the Multi-head self-attention (MHSA) layers to learn new tasks thereby achieving better performance with significantly low memory overhead. We also apply a temporal skip-gating function to tradeoff between stability and plasticity.
    \item Our entropy based approach adds the flexibility of not having to know the task information during inference.
    \item We performed extensive experimentation and ablation of the proposed approach thereby validating the superiority of our model and helping to disentangle the significance of different components.
\end{itemize}

\section{Related Work}
\noindent{\bf Continual Learning:} CL approaches can be broadly classified into --  1)~ Exemplar-replay methods, 2)~regularization methods and 3)~dynamic architecture methods. 
To avoid forgetting when learning a new task, replay approaches repeat past task samples that are kept in raw format~\cite{cvpr2021rm,iclr2019fdr,neurips2020der,icpr2020ert,chaudhry2019continual,douillard2021dytox,aaai2018isele,cvpr2017icarl,wang2022lvt} or generated with a generative model~\cite{neurips2017deepgenreplay}. Usually, replay-based approaches have a fixed memory which stores samples.
Regularization-based approaches, on the other hand, prevent catastrophic forgetting by encouraging important parameters to lie in close vicinity of previous solutions with the introduction of penalty terms to the loss
function ~\cite{PNAS2017kirk,neurips2017lee,zenke2017continual}  or constraining the direction of parameter update~\cite{farajtabar2020orthogonal,iclr2021gpm}. However, the regularization prevents the model from performing well for long task sequences.
Dynamic Architecture methods~\cite{verma2021efficient,cvpr2021der,iclr2018den} grow the model dynamically with the arrival of new tasks. Although these approaches can learn long sequences of tasks, these can suffer from the huge memory and compute overhead if not managed properly.

\begin{figure*}[t!]
\hspace{-4mm}
    \centering
    \includegraphics[scale=0.242]{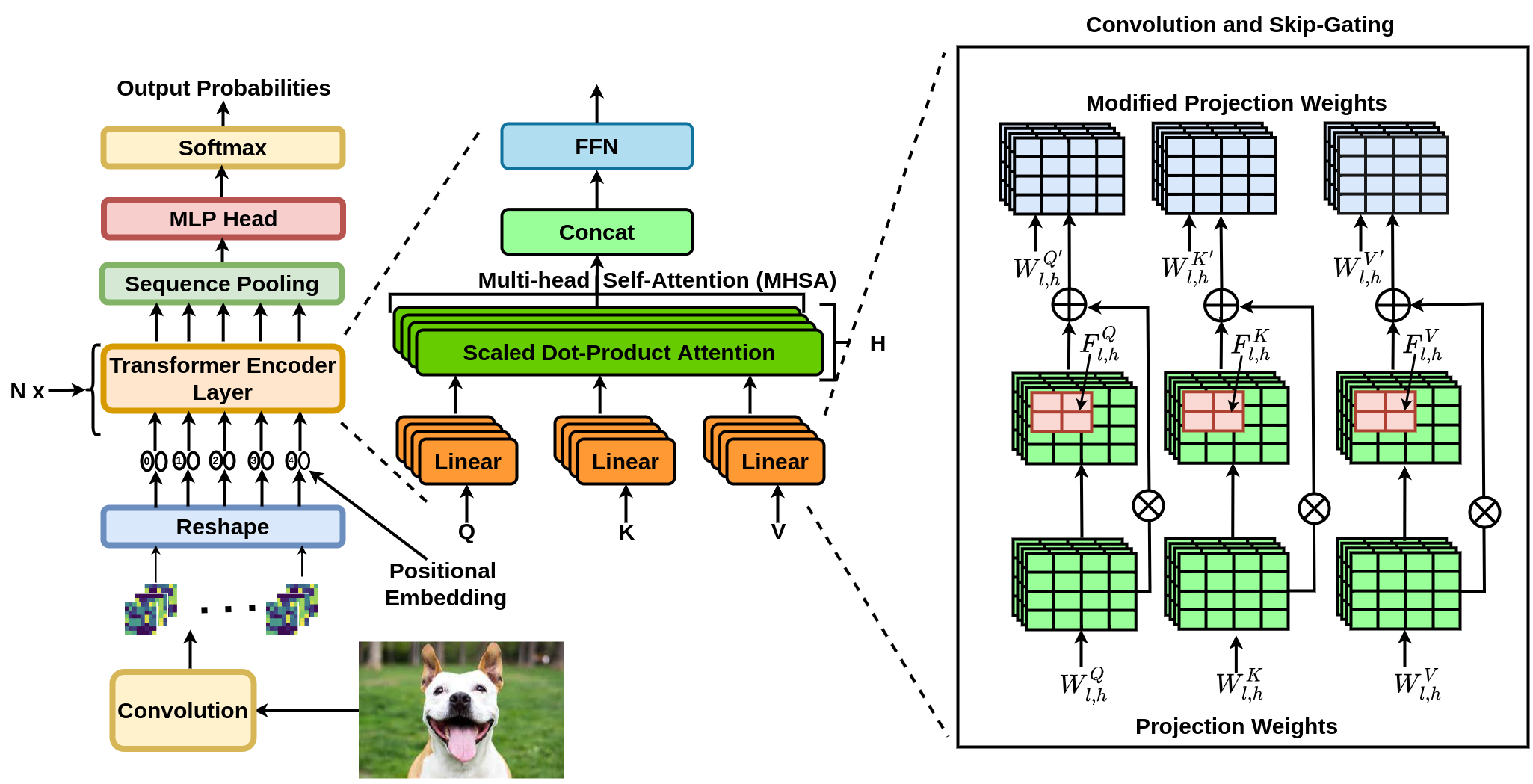}
    \vspace{-3mm}
    \caption{\small Illustration of the proposed \model{} model. {\bf Left} part contains the pictorial representation of the CCT model. {\bf Right} part represents the Convolution and skip-gating operation used to adapt a pre-trained CCT model to new tasks. To learn a new task, the existing set of projection weights (highlighted in \textcolor{forest-green}{\bf green}) are modified by means of adaptable task-specific convolution filters (highlighted in \textcolor{pink}{\bf pink}) and learnable skip-gate (denoted by \pmb{ $\otimes$}), resulting in a fresh collection of projection weights (highlighted in \textcolor{new-blue}{\bf blue}). This eliminates the necessity of developing a new set of projection weights for each new task, and considerably decreases the number of task-specific parameters.}
    \label{fig:transformer_cl}
\vspace{-5mm}
\end{figure*}

\noindent{\bf Transformers:}
Transformers~\cite{neurips2017vaswani} introduced multi-head self-attention (MHSA) in the form of encoder-decoder architecture for machine translation and have since become the state-of-the-art in many NLP tasks~\cite{beltagy2020longformer,brown2020language,choromanski2020rethinking,dehghani2018universal,naacl2019bert}.
In computer vision, the pioneering Vision Transformer (ViT)~\cite{iclr2021vit} directly applied a transformer encoder to image classification with image patches as input.
There have been multiple works since then, including DeiT~\cite{pmlr2021deit}, ConVit~\cite{pmlr2021convit} CCT~\cite{hassani2021cct} \textit{etc.} with architecture and training procedures modifications.

\noindent{\bf Continual Learning in Vision Transformers:} Despite the success of vision transformers, continual learning with vision transformers has received very little attention.
Recently Dytox~\cite{douillard2021dytox} proposed to learn new tasks through the expansion of special tokens known as task tokens.
Another recent approach, LVT~\cite{wang2022lvt}, proposed an inter-task attention mechanism that absorbs the previous tasks’ information and slows down the drift of information between previous and current tasks.
Both Dytox and LVT require extra memory for storing training instances from previous tasks.
Another recent method, MEAT~\cite{cvpr2022meat} uses learnable masks to help isolate previous tasks' parameters that are important for learning new tasks.
However, as the architecture is not expandable, the number of tasks the model can learn is limited.
Additionally, task-ids are required during inference.
A few recent approaches learn soft prompts on new tasks rather than directly modifying the vision encoder which is pre-trained on large amounts of training data.
L2P~\cite{cvpr2022l2p} and DualPrompt~\cite{eccv2022dualprompt} learn a query mechanism to select task relevant prompts from a pool of prompts structured in a key-value scheme.
One drawback of these approaches is they require large pre-training on huge datasets and may fail if the pre-training data is not representative of the later tasks in hand.

Different from the contemporary works, \model{} uses convolution in an intelligent way on the weights that does not require a rehearsal memory for storing samples.
In addition, our proposed approach is flexible enough to operate without the knowledge of task-ids by comparing the entropy of average predictions of several augmented views.
It does not depend on large pre-training and can gracefully handle continual learning scenarios with largely varying tasks.

\section{Method}
In this section, we first describe our problem setting, and then describe the proposed \model{} model. 

\subsection{Problem Setting} 
Considering $T$ tasks arriving sequentially, the goal is to obtain a model which is capable of learning new tasks without forgetting the previous ones.
Specifically, for a task $t \in \{1,2, \ldots, T\}$  the model is exposed to training samples $\{x_i^t , y_i^t\}$, where $x^t_i$ is the $i^{th}$ sample of the $t^{th}$ task and $y^t_i \in C^t$ is the corresponding label belonging to the set of classes $C^t$ with the set of the class labels being mutually exclusive, \textit{i.e.}, $C^0 \cap C^1 \ldots \cap C^T = \phi$.
Our approach is flexible to work in two common CL scenarios, namely (1)~Task Incremental Learning (TIL), and (2)~Class Incremental Learning (CIL). 
In TIL, task-ids for each input is assumed to be known at test time. 
In the more challenging CIL scenario, task-ids are unknown and need to be inferred~\cite{de2021continual,threescenarios2019}.
Our approach is also \textit{exemplar-free}, \textit{i.e.}, data from previous tasks is \textit{not} available for any subsequent use.

\subsection{Preliminaries}
The vision transformer (ViT)~\cite{iclr2021vit} has three major components: (1)~tokenizer layer, (2)~multi-head self-attention (MHSA) and (3)~feed-forward network (FFN).
In ViT, an image is divided into $n$ fixed-size non-overlapping patches.
Each of these patches are embedded into $d$-dimensional vectors.
Formally, the input image $x_i^t$ is tokenized into $n$ patches $z_0 \in \mathbb{R}^{n \times d}$.
Positional embeddings are used to add spatial information in the token sequence.
In ViT, a learnable class embedding is used for the final classification.

A single encoder layer of ViT consists of stacks of MHSA, layer normalization, and FFN blocks, each with residual connections.
At the $l^{th}$ layer, the input is $z_l$ generating the output $z_{l+1}$ that goes into the next layer as input, $l \in \{1, 2, \cdots L\}$ where $L$ is the total number of encoder layers.
Each MHSA block consists of $H$ separate self-attention heads. At the $l^{th}$ layer, self-attention values from head $h \in \{1,2, \cdots H\}$ is:
\begin{align}
    A_{l,h}(Q_{l,h}, K_{l,h}, V_{l,h}) & = softmax \bigg(\frac{Q_{l,h} K^T_{l,h}}{\sqrt{d_k}}\bigg) V_{l,h}
    \label{eq:sa}
\end{align}
where $Q_{l,h}=z_l W^Q_{l,h}$,  $K_{l,h} = z_l W^K_{l,h}$ and $V_{l,h} = z_l W^V_{l,h}$ are query, key, and value with learnable weights $W^Q_{l,h}, W^K_{l,h}$ and $W^V_{l,h} \in \mathbb{R}^{d\times d_k}$ respectively with $d_k$ being the dimension of  key, query and value vectors.
$A_{l,h}$ is the attention matrix obtained as dot product between query and key. 
The activations from different attention heads are then concatenated and passed through a linear projection layer as:
\begin{equation}
    \small
    MHSA_{l} (Q, K, V) = Concat ( A_{l,1},  A_{l,2}, \ldots,  A_{l,H}) W^O
\end{equation}
\normalsize
The FFN block consists of two linear layers and an activation function (usually GELU).

\subsection{ConTraCon}
We now describe the different modules of \model{}.
An overview of the approach is illustrated in Fig.~\ref{fig:transformer_cl}.

\noindent {\bf Convolution on Transformer Weights:}
Self-attention in transformers plays a significant role in their improved performance.
The model learns the relationship between all image patches.
The large effective receptive field of the self-attention layers of transformers helps them to learn the visual representations required for vision tasks~\cite{raghu2021do} by exploiting large amounts of labeled training data.
However, in the presence of only a handful of labeled training data, one of the requirements for capturing the new knowledge without overfitting is to employ only a small number of learnable parameters.
We propose task-specific convolution operations to reweigh the previously learned key, query, and value weights.
The sharable computation offered by the convolutional filters prevents overfitting and maintains plasticity while learning new tasks.
On the other hand, the local computation offered by them helps to maintain the required stability by restricting large changes in the weights. Additionally, convolutional filters being small, help to keep the memory footprint of the newly learned models in check as well.

Specifically, on a ViT, with $L$ encoders and $H$ attention heads per MHSA block, each new task is continually learned by applying convolution kernels on each of the self-attention layers of the MHSA blocks in the transformer:
\begin{equation}
\label{eq:conv_op}
    \begin{split}
        W^{Q'}_{l,h} & = Conv (F^Q_{l,h}, W^Q_{l,h}) \\
        W^{K'}_{l,h} & = Conv (F^K_{l,h}, W^K_{l,h}) \\
        W^{V'}_{l,h} & = Conv (F^V_{l,h}, W^V_{l,h}) \\
    \end{split}
\end{equation}
where $Conv(.)$ denotes the convolution operation which is applied on the weights $W^Q_{l,h}$, $W^K_{l,h}$ and $W^V_{l,h}$ with filters $F^Q_{l,h}, F^K_{l,h}$ and $F^V_{l,h}$ respectively.
Note that the filter weights are learned separately for each task $t \in \{1,2, \ldots, T\}$.
However, to avoid clutter, we are not incorporating the task index above.
These cheap operations help the transformers learn a similar task without forgetting previous tasks and simultaneously keep the compute and memory footprint low.

\noindent {\bf Learnable Skip-Gate:}
It has been shown in ResCL~\cite{Lee2020ResidualCL} that a residual connection combining previous and newly learned weights could improve continual learning performance.
Taking inspiration, we took a step further and replaced the heuristic combination with a learnable skip connection where the combined weight is learned directly from data.
Specifically, we apply the skip connections as follows.
\begin{equation}
    W^{Q''}_{l,h} = W^{Q'}_{l,h} + \sigma (\alpha_{l,h}) * W^Q_{l,h}
\end{equation}
Where $\alpha_{l,h} \in \mathbb{R}$ is a learnable parameter and $\sigma (.)$  is the sigmoid function.
$W^{Q''}_{l,h}$ is the weight matrix of the query for layer 
$l$ and head $h$ corresponding to the new task. Here also, we drop the task index $t$ to avoid clutter.
A similar operation is performed on the key and values weight matrices. 

\subsection{Task Prediction}
In the class incremental setup (CIL), the task-id is not known during inference. Therefore we require a model to predict the task-id of the input in order to decide which set of task-specific parameters to use for inference. This prediction can be difficult, especially when there is no training data available from previous tasks.
One simple approach to predict the task-id is to pass the test image via all task-specific parameters and compare the entropy of all the predictions.
The prediction with the lowest entropy \textit{i.e.}, the prediction with the peakiest confidence can be taken.
However, this method does \textit{not} work well because the cross-entropy training objective tends to make neural networks highly confident, even for out-of-distribution (OOD) samples~\cite{guo2017calibration}. We conjecture that such confident classifiers will predict different classes for different augmentations of an OOD data sample, but their predictions will be consistent for in-distribution samples and their augmentations. Based on this premise, the proposed approach calculates the entropy of the average predictions of a set of augmented views of the input image.
The average entropy reflects the agreement of the predictions for different views of the same image, and thus a highly confident average prediction \textit{i.e.}, the lowest entropy average prediction, is used to infer the task identity.
Once the task identity is inferred, the corresponding task-specific model is invoked on the unaugmented image.
This simple trick is shown to significantly improve task prediction  as verified by ablation studies (see Section~\ref{sec:ablation}). Furthermore, it does so without adding much overhead, except for the increase in batch size due to the augmentations during inference.

\subsection{Training}
As ViT is extremely data hungry and continual learning by nature is data scarce, we resorted to the compact version of ViT proposed by Hassani \textit{et al.}~\cite{hassani2021cct}.
In particular, we used the Compact Convolutional Transformer (CCT) which uses convolutional blocks in the tokenization step reducing the number of effective patches which in turn reduces the number of parameters used later.
CCT further reduces the number of trainable parameters and improves performance by replacing class tokens with a sequence pooling strategy.
Due to its effectiveness in reducing overfitting in a low data regime, we start with a CCT and train it from scratch on the initial task.
Afterward, for every new incoming task, we learn (1) Task-specific convolution filters on the transformer encoder weights, (2) layernorm layers, (3) the sequence pooling layer, and (4) the final classification layer.

\section{Experiments}

We evaluated \model{} on several datasets in both CIL and TIL settings, comparing with many strong state-of-the-art baseline methods.
We also conducted extensive ablation studies to study the effectiveness of different components in our model.
Additional experimental analyses and details can be found in the Appendix.

\subsection{Baselines} 
Following~\cite{wang2022lvt}, we compare \model{} against several well-established rehearsal-based Continual Learning methods such as
iCARL~\cite{cvpr2017icarl}, FDR~\cite{iclr2019fdr}, DER++~\cite{neurips2020der}, ERT~\cite{icpr2020ert}, RM~\cite{cvpr2021rm}. Besides, we also compare our method with various state-of-the-art methods like
Dytox~\cite{douillard2021dytox}, GPM~\cite{iclr2021gpm},  EFT~\cite{verma2021efficient}, LvT~\cite{wang2022lvt}, and  PASS~\cite{cvpr2021pass}.

Inspired by the recent success of Class Attention in image Transformers (CAiT)~\cite{cvpr2021cait}, Dytox~\cite{douillard2021dytox} proposed task attention-based Dynamic Token Expansion for continual classification using transformers.
LvT ~\cite{wang2022lvt} proposes a continual learning mechanism for vision transformers that utilizes an inter-task attention mechanism to consolidate knowledge from previous tasks and avoid catastrophic forgetting. Both Dytox~\cite{douillard2021dytox} and LvT~\cite{wang2022lvt} include a small memory buffer to store training instances of previous tasks. Following ~\cite{wang2022lvt}, we report the performance of all the rehearsal-based approaches with memory buffer sizes of $200$ and $500$.
GPM~\cite{iclr2021gpm} is a regularization-based approach for TIL and so we report only the TIL values for GPM.
EFT~\cite{verma2021efficient} used task-specific feature-map transformation to convolutional architectures.
Specifically, group-wise and point-wise convolutions were used.
For a fair comparison with a similar number of parameters, we ran experiments on EFT by setting the groupsize of the group-wise convolution to the minimum (\textit{i.e., $0$}) and point-wise convolution depth to $8$.
PASS~\cite{cvpr2021pass} performs continual classification by learning per-class prototypes rather than storing exemplars for replay.
Different augmented versions of these stored prototypes are replayed.

\subsection{Datasets, Setup \& Metrics}
We evaluate the continual learning models on four benchmark datasets as follows. \textbf{(1)~CIFAR-100}~\cite{2009cifar} is composed of 60K images of size  $32\times 32$ belonging to $100$ classes with $600$ images per class ($500$ training images and the rest $100$ testing images).
Following~\cite{wang2022lvt} we divided CIFAR-100 into 5 tasks, 10 tasks and 20 tasks.
\textbf{(2)~TinyImageNet-200/10}~\cite{tinyimagenet} is a subset of the ImageNet dataset containing  100K images of size $64\times 64$ distributed among $200$ classes. Each class has $500$ training, $50$ validation and $50$ test images. In TinyImageNet-200/10, the $200$ classes are divided into $10$ sequential tasks containing $20$ classes each.
\textbf{(3)~ImageNet-100/10}~\cite{cvpr2017icarl} contains 100 randomly chosen classes from ImageNet~\cite{russakovsky2015imagenet} having an average resolution of $469\times 387$. It contains around $120K$ images for training and $5K$ for validation. The $100$ classes are divided into $10$ tasks consisting of $10$ classes each.
\textbf{(4)~5-Datasets}~\cite{eccv2020fived} is composed of  CIFAR-10~\cite{2009cifar}, MNIST~\cite{mnist2010}, SVHN~\cite{svhn2011}, Fashion MNIST~\cite{fashionmnist2017} and notMNIST. Classification of each of these datasets is considered as a task.
Additional statistics about the datasets are summarized in Table~\ref{tab:dataset}.

\noindent {\bf Performance metrics:}
Following~\cite{wang2022lvt}, for Class Incremental Learning (CIL), we report top-1 accuracy over all classes of all tasks after training on the last task is completed.
\setlength{\columnsep}{4pt}
\begin{wraptable}{r}{0.36\textwidth}
  \vspace{-1.1\baselineskip}
  \centering
  \footnotesize
    \begin{tabular}{p{1mm}|p{15mm}|p{8mm}|p{5mm}|p{4mm}|p{4mm}}
    \hline
    \multicolumn{2}{c|}{\bf Dataset} & {\bf Size} & {\bf Train} & {\bf Test} & {\bf Class}  \\ 
    \hline
    \multicolumn{2}{l|}{CIFAR-100} & $32\!\times\!32$ &  50K & 10K & 100 \\ 
    \multicolumn{2}{l|}{TinyImageNet-200} & $64\!\times\!64$  & 100K & 10K & 200 \\
    \multicolumn{2}{l|}{ImageNet-100} & 224$\!$\,x\,$\!$224 & 130K & 5K & 100 \\ \hline
    \multirow{5}{*}{{\rotatebox[origin=c]{90}{5-Datasets}}} & CIFAR-10 & 32 x 32 & 50K & 10K & 10 \\ 
    &  MNIST & 32 x 32 & 60K & 10K & 10 \\ 
    &  SVNH & 32 x 32 & 73K & 26K & 10 \\
    & FashionMNIST & 32 x 32 & 60K & 10K & 10 \\
    & notMNIST & 32 x 32 & 60K & 10K & 10 \\ \hline
    \end{tabular}
    \vspace{-2mm}
    \caption{Statistics of the benchmark datasets.}
    \vspace{-1.2\baselineskip}
    \label{tab:dataset}
\end{wraptable}
For Task Incremental Learning (TIL), we report the average accuracy over all the tasks after training on the last task. The average accuracy after training on the $T^{th}$ task is defined by $A_{T} = \frac{1}{T}\sum^{T}_{t=1} a_{T,t}$ where $a_{T,t}$ is the accuracy on the test set of tasks $t$ when the model completed learning task $T$.

\begin{table*}[t!]
    \centering
    \footnotesize
     \addtolength{\tabcolsep}{1.2pt}
    \begin{tabular}{lllllcccccc}
    \hline
        \multirow{2}{*}{\bf Memory Buffer}& \multirow{2}{*}{\bf Model} & \multirow{2}{*}{\bf Approach} & \multirow{2}{*}{\bf Backbone}  & \multirow{2}{*}{\bf \# Params} & \multicolumn{2}{c}{\bf 5 Tasks} & \multicolumn{2}{c}{\bf 10 Tasks} & \multicolumn{2}{c}{\bf 20 Tasks} \\ \cline{6-11}
                & & & & & {\bf TIL} & {\bf CIL} & {\bf TIL} & {\bf CIL} & {\bf TIL} & {\bf CIL} \\ \hline
        \multirow{7}{*}{200} & iCARL~\cite{cvpr2017icarl} & \multirow{7}{*}{Rehearsal}  & \multirow{5}{*}{ResNet 18}  & \multirow{5}{*}{11.2 M} & 55.70 & 30.12 & 60.81 & 22.38 & 62.17 & 12.62 \\ 
        & FDR~\cite{iclr2019fdr} &  &  &  & 63.75 & 22.84 & 65.88 & 14.84 &59.13 & 6.70\\ 
        & DER++~\cite{neurips2020der} &   &  & & 62.55 & 27.46 & 59.54 & 21.76 & 61.98 & 15.16 \\ 
        & ERT~\cite{icpr2020ert} &  & &  & 54.75 & 21.61 & 58.49 & 12.91 & 62.90 & 10.14 \\ 
        & RM~\cite{cvpr2021rm} &   &  & & 62.05 & 32.23 & 66.28 & 22.71 & 68.21 & 15.15 \\ \cline{4-11}
        & LVT~\cite{wang2022lvt} &   & \multirow{2}{*}{Transformer}  & 8.9 M & 66.92 & 39.68 & 72.80 & 35.41 & 73.41 & 20.63 \\ 
        & Dytox~\cite{douillard2021dytox} &   &  & 10.7 M & 75.17 & 40.97 & 84.84 & 32.08 & 85.24 & 15.96 \\ \hline
        \multirow{7}{*}{500} & iCARL~\cite{cvpr2017icarl} & \multirow{7}{*}{Rehearsal}  & \multirow{5}{*}{ResNet 18}  & \multirow{5}{*}{11.2 M} & 64.4 & 35.95 & 71.02 & 30.25 & 72.26 & 20.05\\ 
        & FDR~\cite{iclr2019fdr} &  &   & & 69.11 & 29.99 & 74.22 & 22.81 & 73.22 & 13.10\\ 
        & DER++~\cite{neurips2020der} &   &   &  & 70.74 & 38.39 & 73.31 & 36.15 & 70.55 & 21.65\\
        & ERT~\cite{icpr2020ert} &  &    &  & 62.85 & 28.82 & 68.26 & 23.00 & 73.50 & 18.42 \\ 
        & RM~\cite{cvpr2021rm} &   &   &  & 69.27 & 39.47 & 73.51 & 32.52 & 75.06 & 23.09 \\ \cline{4-11}
        & LVT~\cite{wang2022lvt} &   & \multirow{2}{*}{Transformer}   & 8.9 M & 71.54 & 44.73 & 76.78 & 43.51 & 78.15 & 26.75\\ 
        & Dytox~\cite{douillard2021dytox} &  &  & 10.7 M &  76.1 & 57.66 & {\bf 88.72} & {\bf 47.34}& 87.23 & 29.89\\ \hline
        \multirow{4}{*}{\textendash} & EFT~\cite{verma2021efficient} & Dynamic Arch  & ResNet 18  & 4.9 M (32k) & 79.04 & {\bf 49.68} & 83.14 & 40.42 & 76.75 & 19.15\\ 
        & PASS~\cite{cvpr2021pass} & Regulaization & ResNet 18 & 11.2 M & 70.11 & 47.31 & 71.28 & 35.24 & 71.14 & 23.15 \\
        & GPM~\cite{iclr2021gpm} & Regularization  & AlexNet  & 6.7 M & 65.90 & \textendash & 72.54 & \textendash & 77.59 & \textendash \\ \cline{2-11}
        & \model{} & Dynamic Arch & Transformer & {\bf 3.1 M (26k)} & {\bf 79.37} & 48.46 & 85.69 & 41.26 & {\bf 88.94} & {\bf 30.07} \\ \hline
    \end{tabular}
   
    \caption{Classification accuracy on CIFAR-100 dataset. All methods except EFT, PASS, GPM and \model{} use a memory
buffer of 200 or 500 exemplars. Accuracies in both Task Incremental Learning and  Class Incremental Learning setup are reported under the columns TIL and CIL respectively. The best results are in {\bf bold}. \#Params denote the initial number of parameters in the model (in  millions). 
For dynamic architecture based approaches the extra number of trainable parameters required per task is mentioned inside the parenthesis. 
Even with 500 exemplars, Dytox can  outperform exemplar-free \model{} only on CIFAR-100/10. GPM could not be run in CIL setup and hence its accuracy is not reported.}
     \label{tab:cifar-results-200}
     \vspace{-4mm}
\end{table*}

\subsection{Implementation Details}
We used CCT~\cite{hassani2021cct} as our backbone architecture.
For CIFAR-100, ImageNet-100 and TinyImageNet-200, we used 6 transformer encoder layers with 4 attention heads for each.
For 5-Datasets we used 7 transformer encoder layers with 4 attention heads per layer.
The encoder layers have an embedding dimension of 256.
We used sinusoidal positional embeddings.
For CIFAR-100 and 5-Datasets, we configured the tokenizer to a single convolution layer with kernel size $3$.
For TinyImageNet-200/10, the tokenizer consisted of $2$ convolution layers with kernel size $5$ and for ImageNet-100/10 the tokenizer had $3$ convolution layers with kernel size $3$.
A stride of $2$ and padding of $3$ were used for all datasets.
Following CCT~\cite{hassani2021cct}, we set both attention and depth dropout probability to $0.1$.
Each task was trained for $500$ epochs using AdamW~\cite{iclr2019adamw} optimizer.
We employed cosine annealing learning rate scheduler with initial learning rate of $8e^{-4}$ and warm restarts~\cite{loshchilov2016sgdr}.
For the convolution operation on the transformer weights, we set kernel size ($k$) as $15$, which is obtained by validating on a small subset of the data.
Convolution is performed on the transformer weights learnt after the initial task \textit{i.e.}, $W^Q_{l,h}, W^K_{l,h}$ and $W^V_{l,h}$ in Eqn.~\ref{eq:conv_op} comes from the first task.
For task prediction, using 10 augmentations and $\beta=0.6$ gives the best result.

\begin{table*}[t!]
    \centering
    \footnotesize
     \addtolength{\tabcolsep}{3.5pt}
    \begin{tabular}{lllllcccc}
    \hline
        \multirow{2}{*}{\bf Memory Buffer} & \multirow{2}{*}{\bf Model} & \multirow{2}{*}{\bf Approach} & \multirow{2}{*}{\bf Backbone}  &  \multirow{2}{*}{\bf \# Params} & \multicolumn{2}{c}{\bf ImageNet-100/10} & \multicolumn{2}{c}{\bf TinyImageNet-200/10} 
        \\ \cline{6-9}
                & &  & & & {\bf TIL} & {\bf CIL} & {\bf TIL} & {\bf CIL} \\ \hline
       \multirow{7}{*}{200} & iCARL~\cite{cvpr2017icarl} &  & \multirow{5}{*}{ResNet 18}  & \multirow{5}{*}{11.2 M} & 33.75 & 12.59 & 28.41 & 8.64 \\ 
        & FDR~\cite{iclr2019fdr} & \multirow{7}{*}{Rehearsal} & & & 37.80 & 10.08 & 40.15 & 8.77  \\ 
        & DER++~\cite{neurips2020der} & & & & 31.96 & 11.92 & 40.97 & 11.16  \\ 
        & ERT~\cite{icpr2020ert} & & & & 36.94 & 13.51 & 39.54 & 10.85  \\ 
        & RM~\cite{cvpr2021rm} & & & & 35.18 & 16.76 & 41.96 & 13.58  \\ \cline{4-9}
        & LVT~\cite{wang2022lvt} & & \multirow{2}{*}{Transformer}   & 9.0 M & 41.78 & 19.46 & 46.15 & 17.34 \\ 
        & Dytox~\cite{douillard2021dytox} & & & 10.7 M & 70.12 & 41.76 & 61.71 & 19.14\\ \hline
        \multirow{7}{*}{500} & iCARL~\cite{cvpr2017icarl} & \multirow{7}{*}{Rehearsal} & \multirow{5}{*}{ResNet 18}  & \multirow{5}{*}{11.2 M} & 36.89 & 16.44 & 35.89 & 10.69  \\ 
        & FDR~\cite{iclr2019fdr} & & & & 42.60 & 11.78 & 49.91 & 10.58 \\ 
        & DER++~\cite{neurips2020der} & & & & 35.46 & 14.52 & 51.90 & 19.33 \\ 
        & ERT~\cite{icpr2020ert} & & & & 41.56 & 20.42 & 50.87 & 12.13  \\ 
        & RM~\cite{cvpr2021rm} & & & & 38.66 & 14.56 & 52.08 & 18.96  \\ \cline{4-9}
        & LVT~\cite{wang2022lvt} & & \multirow{2}{*}{Transformer} & 9.0 M & 47.84 & 26.32 & 57.93 & 23.97 \\ 
        & Dytox~\cite{douillard2021dytox} & & & 10.7 M & 73.64 & 40.94 & {\bf 64.29} & 26.39 \\ \hline
        \multirow{4}{*}{\textendash} & EFT~\cite{verma2021efficient} & Dynamic Arch  & ResNet 18 & 4.9 M (32k) & 72.18 & 32.98 & 60.00 & 24.08 \\ 
        & PASS~\cite{cvpr2021pass} & Regularization & ResNet 18 & 11.2 M & 39.9 & 34.52 & 43.9 & 22.76 \\
        & GPM~\cite{iclr2021gpm} & Regularization  & AlexNet  & 6.7 M & 40.65 & \textendash & 45.48 & \textendash \\ \cline{2-9}
        & \model{} & Dynamic Arch & Transformer & {\bf 3.6 M (28k)} & {\bf 76.78} & {\bf 42.2} & 62.76 & {\bf 27.46} \\ \hline
    \end{tabular}
   
    \caption{Classification accuracy on ImageNet-100/10, TinyImageNet-200/10. All methods except EFT, PASS, GPM and \model{} use a memory
buffer of 200 or 500 exemplars. 
The number of extra parameters required per task is mentioned in brackets for dynamic architecture based approaches. 
\model{} outperforms all the baselines with memory buffers 200 and 500.
For \model{} the number of parameters for ImageNet-100/10 and TinyImageNet/10 are different. We mentioned the greater of the two in the table.
GPM could not be run in CIL setup and hence its accuracy is not reported.}
\label{tab:imgnet-tiny-results-200}
\vspace{-5mm}
\end{table*}

\begin{table}[t!]
    \centering
    \addtolength{\tabcolsep}{-3pt}
    \footnotesize
    \begin{tabular}{llllll}
    \hline
        \multirow{2}{*}{\bf Model} & \multirow{2}{*}{\bf Approach} & \multirow{2}{*}{\bf Backbone}  &  \multirow{2}{*}{\bf \# Params} & \multicolumn{2}{c}{\bf 5-Datasets} \\ \cline{5-6}
                & &  & & {\bf TIL} & {\bf CIL} \\ \hline
        Dytox~\cite{douillard2021dytox}(500) & Rehearsal  & Transformer  & 10.7 M &  77.12 & {\bf 67.13} \\ \hline
        Dytox~\cite{douillard2021dytox}(200) & Rehearsal  & Transformer  & 10.7 M &  75.81 & 65.04 \\ \hline
        EFT~\cite{verma2021efficient} & Dynamic Arch  & ResNet 18 & 4.9 M (32k) & 94.75 & 52.04\\ 
        GPM~\cite{iclr2021gpm} & Regularization  & ResNet18  & {\bf 1.2 M} & 90.60 & \textendash \\ \hline
        \model{} & Dynamic Arch & Transformer &  3.9 M (28k) &  {\bf 95.10} & 65.21\\ \hline
    \end{tabular}   
    \caption{ Classification accuracy on 5-Datasets. Mentioned within parentheses are the number of additional parameters required to learn each new task for dynamic architecture based approaches like EFT and \model{}. The memory buffer sizes for Dytox are mentioned inside the parenthesis.}
     \label{tab:5d-results}
\end{table}

\subsection{Results and Analyses}
Tables~\ref{tab:cifar-results-200}, \ref{tab:imgnet-tiny-results-200} and~\ref{tab:5d-results} compare \model{} with baselines working on both CNN and transformers, over the aforementioned four datasets.
The results show the performance of our approach in both CIL and TIL setups.
Table~\ref{tab:cifar-results-200} shows the results on the CIFAR-100 dataset split into 5 tasks, 10 tasks, and 20 tasks following the standards commonly adopted in the community.
The 5-tasks split contains fewer tasks but more classes per task.
The 20-tasks split is more challenging, where performance of most models suffer a drastic drop compared to setups consisting of fewer tasks.
\model{} significantly outperforms  existing approaches over longer sequences of tasks.
In task incremental setting (TIL), \model{} performs the best across all the splits using the lowest amount of parameters.
As rehearsal and regularization based approaches do not expand their architectures with new tasks, we provide the number of initial parameters learned for dynamic architecture expansion based approaches including ours, for fair comparison (ref \textbf{\# Params} column).
We also provide the number of additional parameters per task for the dynamic expansion based approaches including ours.

For the challenging CIL setup, \model{} does the best in the most challenging 20-Tasks scenario.
For 10-Tasks, \model{} is the best among the dynamic architecture based approaches and better than most rehearsal based approaches.
An advantage of our approach is that it does not need an extra memory buffer to store previous examples while remaining the most parameter efficient.

Table~\ref{tab:imgnet-tiny-results-200} shows the performance on Imagenet-100/10 and TinyImagenet-100/10.
The format \emph{dataset-X/Y} implies that the dataset contains a total of X classes divided into Y tasks uniformly.
\model{} significantly beats all the competing approaches while remaining parameter efficient and without using replay memory.
Table~\ref{tab:5d-results} shows the performance on the challenging 5-Datasets.
While requiring more parameters compared to GPM, \model{} significantly outperforms it in the TIL setup.
\model{} can equally handle CIL setup which GPM can not.
In the challenging CIL setup, DyTox is better but at the cost of using almost 2.5 times the number of parameters as used in \model{}, and using 500 exemplars (while \model{} is exemplar-free).

\subsection{Model Introspection}

In this section, we analyze and explain the significance of various components of our approach.
For this purpose, we use the CIFAR-100/10 dataset unless otherwise mentioned.

\begin{figure*}[!t]
\captionsetup[subfigure]{aboveskip=-1pt,belowskip=-1pt}
\input{figs/kernel_size_ablation}
\input{figs/split_ratio}
\input{figs/with_and_without_aug}
\input{figs/aug_vs_time_plot}
\caption{Ablation studies on \model{} over CIFAR-100/10 dataset: (a)~Average classification accuracy using different convolutional kernel sizes. (b)~Effect of the presence of local filters -- interestingly, the presence of task adaptable convolution filters in the tokenizer layer lowers the performance. 
(c)~Effect of augmentation in the inference. Reported is the average top-1 accuracy in the CIL setup calculated after training all the tasks. \model{}'s performance drops significantly when inference is done without augmentation. (d) Inference time vs number of augmentations for one batch of $64$ images.}
\vspace{-1\baselineskip}
\label{fig:ablation}
\end{figure*}

\noindent{\bf Upper and Lower Bounds:}
We trained separate CCT backbones for each task assuming no limitations on the total number of learnable parameters.
The average accuracy was found to be $89.3\%$ which serves as the performance upperbound having $100\%$ increase in parameters for each new task.
On the other extreme, we fine-tuned an initial CCT backbone with every new task giving 0 parameter increase per task but suffering from catastrophic forgetting the most.
The average accuracy (calculated after training the last task) obtained is only $17.1\%$.
\model{} achieves $85.7\%$ average classification accuracy in TIL setup, a mere $\sim 4\%$ drop in comparison to the upper bound while using only $0.7\%$ of the parameters required per task by the upper bound.

\begin{figure}[H]
\vspace{-2mm}
    \includegraphics[width=0.95\linewidth]{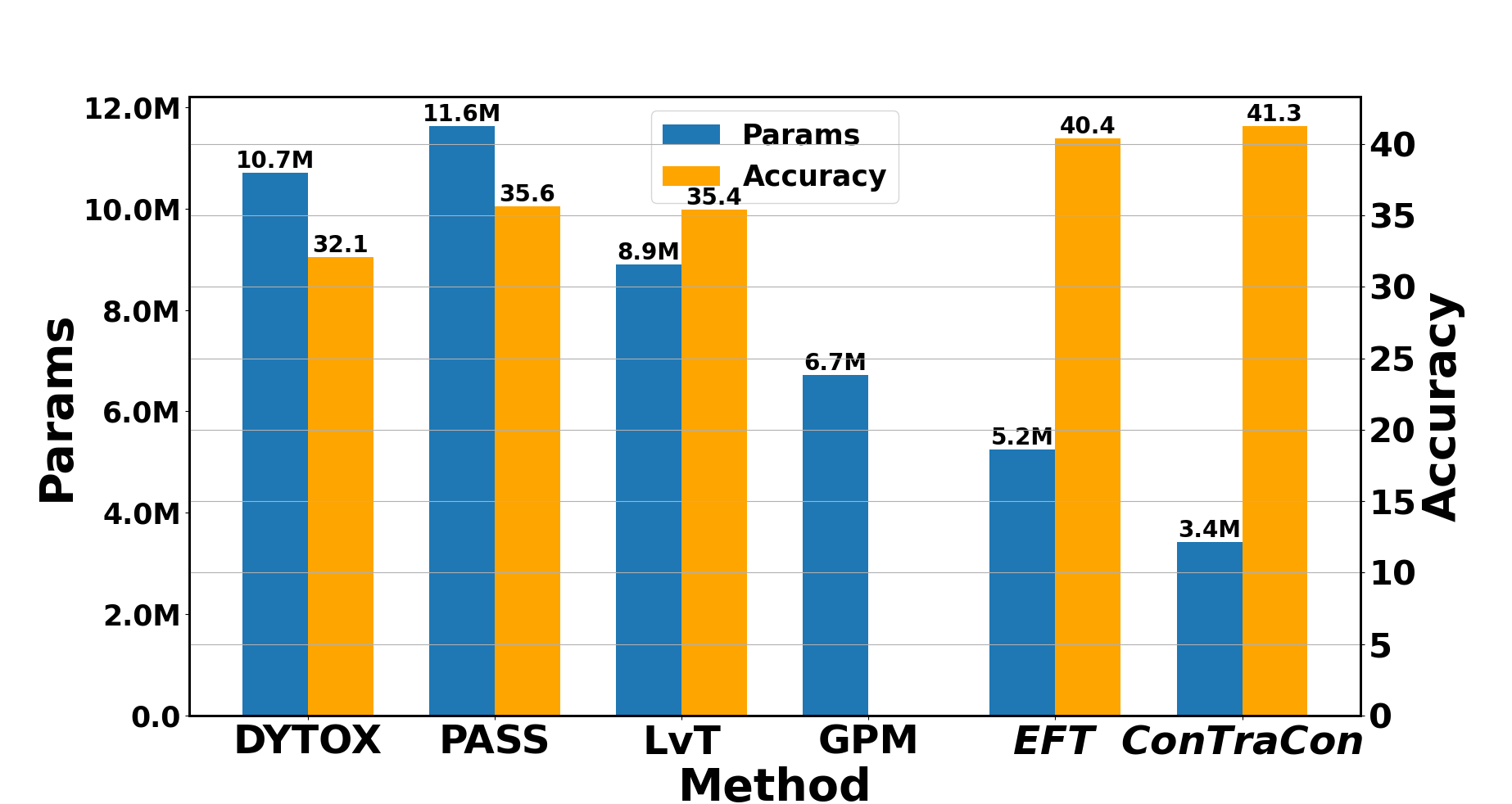}
    \caption{When models are learning CIFAR-100/10 tasks in CIL setup (with 200 buffer size for exampler-replay approaches): \textcolor{blue}{blue} bars -- the total number of parameters required (left $y$ axis), \textcolor{orange}{orange} bars -- the average classification accuracy (right $y$ axis). 
    GPM could not be run in CIL setup and hence its accuracy is not reported.  
    \model{} uses significantly less parameters, yet achieves higher accuracy.}
    \label{fig:parameter_number}
    \vspace{-2mm}
\end{figure}
\noindent{\bf Memory Overhead: }
Fig.~\ref{fig:parameter_number} shows the number of trainable parameters required in CIFAR-100/10 dataset. As can be seen, \model{} has the least number of trainable parameters. 
For each novel task, we have separate convolution kernels, layernorm, sequence pool, and classification layers. These are cheap to store as they consist of very few parameters. \model{} incurs a total parameter increase of $\sim 7\%$ for learning all the tasks in CIFAR-100/10.

\noindent{\bf Computational Overhead:} \model{} uses $50\%$ less total FLOPs during training compared to DyTox.
This is because Dytox uses additional distillation operation requiring the inputs to be passed through the transformer twice, while \model{} uses cheap convolution operation on transformer weights.
For predicting the task during inference, we compute the entropy of average predictions of a number of augmentations of the input image.
While augmentation-based inference can principally incur more computation, due to highly parallel frameworks, the overhead is less and scales linearly with number of augmentations.
We ran a small experiment on CIFAR-100 with batch-size 64 with augmentations ranging from $5$ to $30$ in steps of $5$ (Fig.~\ref{fig:ablation}(d)).
Going from no augmentation to $10$, inference time changes from $0.14$s to $0.2$s \textit{i.e.}, an overhead of only $94$ms per image.
With the performance gain (Fig.~\ref{fig:ablation}(c)), our augmentation based strategy is lightweight for all practical purposes.

\subsection{Ablation Study}
\label{sec:ablation}

We take a closer look at our method with ablation studies on the CIFAR-100/10 dataset as described below.

\noindent{\bf Effect of Convolution Kernel size:} We varied the kernel size from 3 to 23 (in steps of 2) and noted the average performance.
Fig.~\ref{fig:ablation}(a) shows that the performance increases with an increase in the kernel size until the best tradeoff is reached roughly at a kernel size of 15.
A further increase of the kernel size merely improves the performance by $0.2\%$ ($85.7\%$ vis-a-vis $85.9\%$) but at the cost of a parameter increase of $47k$ per task compared to $26k$ with kernel size $15$.

\noindent{\bf Effect of task adaptable convolutions in Tokenizer Layer:}
We wanted to see the effect of task adaptable convolution on the tokenizer layer.
To this end, we divided the total number of filters in the tokenizer into global filters and local filters.
The global filters are frozen after they are trained once while the local filters are trained specific to each  task.
the ratio of the number of global filters to total filters is termed as the {\it split-ratio}.
A high split-ratio implies less task-specific filters whereas a low split-ratio indicates a higher number of task-specific filters and thus a higher number of task-specific parameters.
A lower split ratio may adapt better to newer tasks as more parameters are dedicated to task specific learning.
However as seen in Fig.~\ref{fig:ablation}(b), this is not the case.
We conjecture that higher number of task adaptable parameters at the stem of the network can destroy the already learnt knowledge which may be crucial at the tokenizer layer of the CCT backbone.
As, highest split ratio (all global filters) gives the highest accuracy, we conducted all the experiments with the highest split ratio.

\noindent{\bf Effect of Augmentation:} To better understand the importance of input image augmentation in task prediction, we performed a comparative analysis.
As shown in Fig.~\ref{fig:ablation}(c),
a significant drop in CIL performance can be witnessed when the task prediction is done without augmentation. This is because models trained with softmax cross-entropy loss are over-confident in their prediction~\cite{guo2017calibration}, even if the instance does not belong to the task it is trained for.
So we used average predictions of the augmentations for task prediction.

\noindent{\bf Effect of Skip-Gating:} In order to guaze the effect of skip-gating
we ran three experiments -- one without any skip connection ($\sigma(\alpha)=0$), one with a learnable skip gating and one with skip connection but no gating ($\sigma(\alpha)=1$).
We found the one without the skip connection got an average accuracy of $83.5\%$, the one with a skip connection but without gating had an average performance of $84.7\%$ whereas the one with learnable skip-gating got an accuracy of $85.7\%$ showing the importance of skip-gating for \model.

\section{Conclusion}

We propose Continual Transformer with Convolutions (\model{}), an approach that is capable of learning incoming tasks efficiently (through low-cost operations), allowing the model to expand its knowledge without blowing up its size and computation, and without having to store data samples for rehearsal.
We validate the superiority of \model{} through extensive experimentation.
We also perform thorough ablation studies highlighting the significance of various components of our approach.
To our knowledge, this is one of the early works on making transformers adaptable to continual learning, and we hope our demonstration of performance and scalability will drive further research in making transformers adaptable to continual learning.

{\small
\bibliographystyle{ieee_fullname}
\bibliography{contracon}
}

\section*{Appendix}
\appendix

This appendix contains the following.
\begin{itemize}[leftmargin=*]
	\item Appendix~\ref{sec:upper_bound} Upper Bounds on classification accuracies on various datasets.
	\item Appendix~\ref{sec:5d_res}: Classification Accuracies on 5-Datasets over more baselines.
	\item Appendix~\ref{sec:rev_task_res} Classification Accuracies of \model{} with different task orders.
	\item Appendix~\ref{sec:aug}: Augmentations used for task prediction
\end{itemize}

\section{Upper Bounds on Classification Accuracies}
\label{sec:upper_bound}

To better understand the performance of \model{}, we calculate the upper bounds, i.e., the maximum achievable performance by the backbone architecture. 
Specifically, we train each task on a separate backbone architecture, thereby having a per-task parameter increase of $100\%$. Using this setup, we calculate the upper-bounds for all the datasets and task-splits. Table~\ref{tab:upper_bound} shows a comparative study of the performance variation between the upper-bound and \model{}.
On average, we observe that \model{}'s performance is $\sim 1-4 \%$ below the corresponding upper-bounds while requiring only $0.7\%$ of the number of parameters required per task for the upper-bound performances.
\begin{table}[ht]
    \centering
    \footnotesize
    \begin{tabular}{p{27mm}llll}
    \hline
        \multirow{2}{*}{\bf Dataset} & \multicolumn{2}{c}{\bf Upper Bound} &  \multicolumn{2}{c}{\bf \model{}} \\ \cline{2-3}\cline{4-5}
                & {\bf TIL} & {\bf \# Params} & {\bf TIL} & {\bf \# Params}  \\ \hline

        CIFAR-100/5  & 85 & 3.1 M & 79.37 & 26k \\ 
        CIFAR-100/10  & 89.30 & 3.1 M & 85.96 & 26k \\
        CIFAR-100/20  & 93.16 & 3.1 M & 88.94 & 26k \\ 
        ImageNet-100/10 & 80.67 & 3.6 M &  76.78 & 28k \\
        TinyImageNet-200/10 & 70.66 & 3.6 M & 62.76 & 28k \\
        5-Datasets & 96.42 & 3.9 M & 95.10 & 28k \\ \hline
\end{tabular}
    \caption{Performance comparison of \model{} with the upper-bound calculated by training the backbone for each task. TIL values denote the average classification accuracy in the Task Incremental Learning setup.
    \# Params denotes the number of parameters required to learn each task.}
     \label{tab:upper_bound}
\end{table}
\section{Results on 5-Datasets}
\label{sec:5d_res}

\begin{figure*}[!t]
\begin{subfigure}{0.2\textwidth}
    \includegraphics[width=0.95\linewidth]{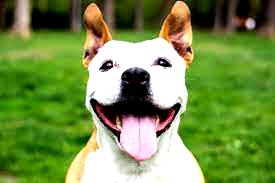}
    \caption{}

    \label{fig:aug1}
\end{subfigure}%
\begin{subfigure}{0.2\textwidth}
    \includegraphics[width=0.95\linewidth]{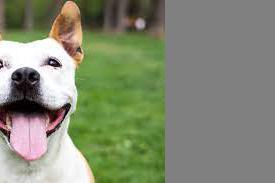}

\caption{}
    \label{fig:aug2}
\end{subfigure}%
\begin{subfigure}{0.2\textwidth}
    \includegraphics[width=0.95\linewidth]{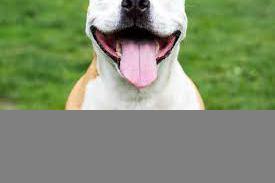}

\caption{}
    \label{fig:aug3}
\end{subfigure}%
\begin{subfigure}{0.2\textwidth}
    \includegraphics[width=0.95\linewidth]{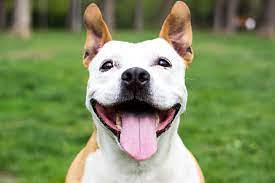}
\caption{}
    \label{fig:aug4}
\end{subfigure}%
\begin{subfigure}{0.2\textwidth}

    \includegraphics[width=0.95\linewidth]{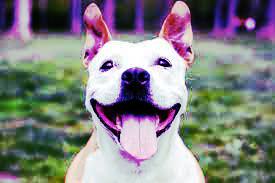}
\caption{}
    \label{fig:aug5}
\end{subfigure}
\begin{subfigure}{0.2\textwidth}

    \includegraphics[width=0.95\linewidth]{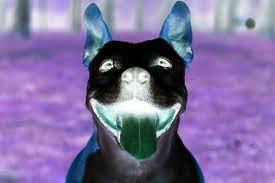}
\caption{}
    \label{fig:aug6}
\end{subfigure}%
\begin{subfigure}{0.2\textwidth}

    \includegraphics[width=0.95\linewidth]{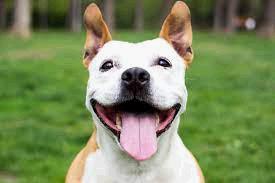}

\caption{}
    \label{fig:aug7}
\end{subfigure}%
\begin{subfigure}{0.2\textwidth}

    \includegraphics[width=0.95\linewidth]{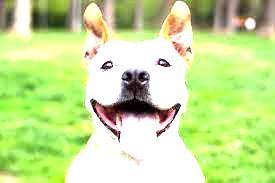}
\caption{}
    \label{fig:aug8}
\end{subfigure}%
\begin{subfigure}{0.2\textwidth}
    \includegraphics[width=0.95\linewidth]{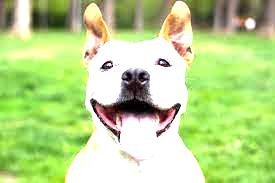}

\caption{}
    \label{fig:aug9}
\end{subfigure}%
\begin{subfigure}{0.2\textwidth}
    \includegraphics[width=0.95\linewidth]{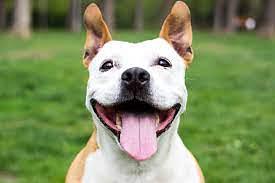}
\caption{}
    \label{fig:aug10}
\end{subfigure}%
\vspace{-4mm}
\caption{Augmentations used for Task-id prediction (a) increased contrast (b) translation along x axis (c) Translation along y-axis (d) increased sharpness (e) Equalized image (f) Inverted image (g) posterized image (h) increased brightness (i) increased brightness (j) increased sharpness. }

\end{figure*}

First proposed by Ebrahimi \textit{et al.}~\cite{eccv2020fived}, 5-Datasets
is composed of CIFAR-10~\cite{2009cifar},
MNIST~\cite{mnist2010}, SVHN~\cite{svhn2011}, Fashion MNIST~\cite{fashionmnist2017} and notMNIST where classification on each of these datasets is a task. 
The variation / diversity in the dataset for each of the tasks in 5-Dataset sets it apart from the other benchmark datasets used in this paper.
As the dataset is very recently proposed, some of the competitive continual learning approaches did not have chance to validate on this dataset.
Hence, we ran a few recent approaches on this dataset and compared with \model{}.
Specifically, we ran DER++~\cite{neurips2020der} and FDR~\cite{iclr2019fdr} on this very challenging dataset.

Table~\ref{tab:5d-result} shows the results on this dataset.
While FDR~\cite{iclr2019fdr} and DER++~\cite{neurips2020der} use ResNet18 as the backbone, GPM and EFT uses a variation of it to reduce the number of learnable parameters with an eye to avoid possible overfitting.
It can be noted that \model{} significantly beats FDR and DER++ on this challenging and diverse continual learning dataset with almost 15\% gain over the best of the two in (Task Incremental Learning) TIL and around 20\% better in the (Class Incremental Learning) CIL setting.
Additionally \model{} uses much less parameters (almost 33\% less) compared to both FDR and DER++ showing the capability of our proposed task adaptable convolution to handle diverse tasks for continual learning in TIL as well as CIL settings.
\begin{table}[ht!]
    \centering
    \footnotesize
    \begin{tabular}{p{12mm}p{11mm}p{11mm}p{13mm}p{5mm}p{5mm}}
    \hline
        \multirow{2}{*}{\bf Model} & \multirow{2}{*}{\bf Approach} & \multirow{2}{*}{\bf Backbone}  &  \multirow{2}{*}{\bf \# Params} & \multicolumn{2}{c}{\bf 5-Datasets} \\ \cline{5-6}
                & &  & & {\bf TIL} & {\bf CIL} \\ \hline

        FDR~\cite{iclr2019fdr} & Rehearsal & ResNet18 & 11.2 M & 72.45 & 38.21\\ 
        DER++~\cite{neurips2020der} & Rehearsal & ResNet18 & 11.2 M & 80.45 & 45.03\\ 
        \hline
        EFT~\cite{verma2021efficient} & Dyn Arch  & ResNet18 & 4.9 M (32k) & 94.75 & 52.04\\ 
        GPM~\cite{iclr2021gpm} & Reg  & ResNet18  & {\bf 1.2 M} & 90.60 &  \textendash \\
        Dytox~\cite{douillard2021dytox} & Rehearsal  & Transformer  & 10.7 M & 77.12 & {\bf 67.13} \\ \hline
        \model{} & Dyn Arch & Transformer &  3.9 M (28k) &  {\bf 95.10} & 65.21\\ \hline
\end{tabular}
    \caption{Classification accuracy on 5-Datasets. Additional parameters per task for dynamic architecture-based approaches are mentioned inside brackets.. EFT and GPM use a reduced version of the resnet18 architecture as their backbone. The rehearsal based approaches use a buffer of size $500$.}
     \label{tab:5d-result}
\end{table}
\section{Results with Different Task Orders}
\label{sec:rev_task_res}

\model{} uses convolution-based task adaptation over the original backbone (CCT~\cite{hassani2021cct}), pre-trained on the first task.
To show the robustness of our approach on the choice of the initial task, we chose to experiment with different initial tasks out of the tasks available for CIFAR-100/10.

For this purpose, we perform two variations of the experiments shown in Table~\ref{tab:cifar-results-200} -- 
(1)~Train \model{} with \textit{reversed} task order as followed in Table~\ref{tab:cifar-results-200} so that the initial task there becomes the final task here and vice-versa, and
(2)~Train \model{} with a random task-order.

 We observe that, with task-order reversed, the model achieves an average classification accuracy of $84.83\%$ in the TIL setup while the same accuracy for the original task-order followed in Table 2 of the main paper is $85.69\%$.
 With random task-order, \model{} achieves an average classification accuracy of $83.82\%$.
 Note that, even with different task-orders, \model{}'s performance is always better than that of the state-of-the-art approaches.
\section{Augmentations Used}
\label{sec:aug}

Entropy-based task prediction performs poorly due to
\begin{wrapfigure}{r}{0.2\textwidth}
	\vspace{-1.6\baselineskip}
  \begin{center}
    \includegraphics[width=0.2\textwidth]{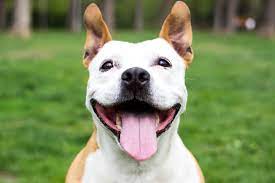}
  \end{center}
	\vspace{-4mm}
  \caption{Original image}
  \label{fig:dog}
  \vspace{-1.5\baselineskip}
\end{wrapfigure}
cross-entropy training loss function.
This results in high confidence (i.e., low entropy) predictions even for out-of-distribution inputs~\cite{guo2017calibration}. Hence, to overcome this, we calculate the  entropy of the average predictions of different augmentations of the input image (as discussed in Section 3.4 of the main paper).

Specifically, for an input image during test time (shown in Fig.~\ref{fig:dog}), we augment the test image in 10 different ways.
After that we pass these through various task specific models and calculate the entropy of each of the predicted probability distributions to ultimately get the task ids.
In this appendix, we provide the details of the augmentations we used for this purpose. The augmented versions of the image in Fig.~\ref{fig:dog} are shown in Fig.~\ref{fig:aug1}--Fig.~\ref{fig:aug10}.

\end{document}